\documentclass{article}

\usepackage{arxiv}

\usepackage{doi}

\usepackage[utf8]{inputenc} 
\usepackage[T1]{fontenc}    
\usepackage{hyperref}       
\usepackage{url}            
\usepackage{booktabs}       
\usepackage{amsfonts}       
\usepackage{amsmath}

\usepackage{todonotes}
\usepackage{caption}
\usepackage{subcaption}
\usepackage{cancel}

\usepackage{nicefrac}       
\usepackage{microtype}      
\usepackage{lipsum}
\usepackage{fancyhdr}       
\usepackage{graphicx}       
\graphicspath{{media/}}     

\date{}
  
\title{A connection between probability, physics \\and neural networks
\thanks{\textit{\underline{Citation}}: 
\textbf{Ranftl, S. A connection between probability, physics and neural networks. Proceedings of MaxEnt 2022.  }} 
}

\author{ \href{https://orcid.org/0000-0001-8956-2576}{\includegraphics[scale=0.06]{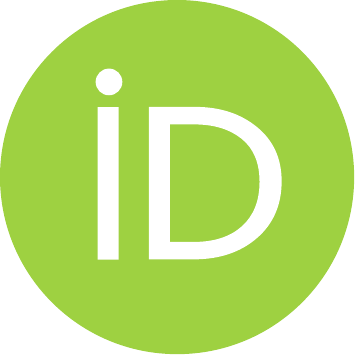}\hspace{1mm}Sascha Ranftl}  \\
	Institute of Theoretical Physics - Computational Physics\\
	Graz University of Technology\\
	Graz, Austria \\
	\texttt{ranftl@tugraz.at} \\
}

\renewcommand{\shorttitle}{A connection between probability, physics and neural networks}
\renewcommand{\undertitle}{}

\hypersetup{
pdftitle={A connection between probability, physics and neural networks},
pdfsubject={stat.ML, physics.data-an, math.PR, cs.AI},
pdfauthor={Sascha Ranftl},
pdfkeywords={Bayes, probability, neural networks, Gaussian process, kernels, covariance functions, activation functions, physics-informed machine learning, differential equations, PDEs, ODEs, linear operators, linear constraints, inverse kernel trick},
}

\begin{document}
\maketitle

\begin{abstract}
We illustrate an approach that can be exploited for constructing neural networks which  a priori obey physical laws.
We start with a simple single-layer neural network (NN) but refrain from choosing the activation functions yet. Under certain conditions and in the infinite-width limit, we may apply the central limit theorem, upon which the NN output becomes Gaussian. We may then investigate and manipulate the limit network by falling back on Gaussian process (GP) theory. It is observed that linear operators acting upon a GP again yield a GP. This also holds true for differential operators defining differential equations and describing physical laws. If we demand the GP, or equivalently the limit network, to obey the physical law, then this yields an equation for the covariance function or kernel of the GP, whose solution equivalently constrains the model to obey the physical law. The central limit theorem then suggests that NNs can be constructed to obey a physical law by choosing the activation functions such that they match a particular kernel in the infinite-width limit. The activation functions constructed in this way guarantee the NN to a priori obey the physics, up to the approximation error of non-infinite network width. Simple examples of the homogeneous 1D-Helmholtz equation are discussed and compared to naive kernels and activations. 
\end{abstract}

\keywords{Bayes\and probability\and neural networks\and Gaussian process\and kernels\and covariance functions\and activation functions\and physics-informed machine learning\and differential equations\and PDEs\and ODEs\and linear operators\and linear constraints\and inverse kernel trick}

\definecolor{Gray}{gray}{0.9}
\LetLtxMacro{\originaleqref}{\eqref}
\newcommand\figref{fig.~\ref}
\newcommand\sectref{sect.~\ref}
\newcommand\tabref{table~\ref}

\newcommand{\expect}[1]{\langle #1  \rangle}
\newcommand{\condexpect}[2]{\langle #1 \mid #2 \rangle}
\newcommand{\BK}{\mathcal{I}}
\newcommand{\p}[2]{{p}\big(#1   \boldsymbol{\mid}  #2 \big)}
\newcommand{\prior}[1]{{p} \big(#1\big)}
\newcommand{\pN}[2]{\mathcal{N}\Big[#1 {\color{red} \boldsymbol{\mid}} #2 \Big]}
\newcommand{\pr}{^{\prime}}
\renewcommand{\a}{\boldsymbol{a}}
\renewcommand{\d}{\boldsymbol{{d}}}
\newcommand{\gp}{\tilde{z}}
\renewcommand{\b}{\boldsymbol{\beta}}
\newcommand{\de}{\d_{exp}}
\newcommand{\ds}{\d_{sim}}
\newcommand{\hparams}{c}
\newcommand{\vhparams}{\boldsymbol{\hparams}}
\newcommand{\z}{z}
\newcommand{\boldz}{\boldsymbol{\z}}
\newcommand{\pq}{\boldsymbol{p}}
 
\newcommand{\etav}{\boldsymbol{\eta}}
\newcommand{\gamdotv}{\boldsymbol{\dot \gamma}}
\renewcommand{\a}{\boldsymbol{a}}
\newcommand{\dgamma}{\dot{\gamma}}
\newcommand{\x}{\dgamma}
\newcommand{\vv}[1]{{\boldsymbol #1}}
\newcommand{\avg}[1]{\left\langle #1 \right \rangle}
\newcommand{\opttxt}[1]{\textcolor{blue}{#1}}
\newcommand{\soutw}[1]{\sout{\textcolor{white}{#1}}}

\newcommand{\dexp}{\vv d_\text{exp}}
\newcommand{\dsim}{\vv d_\text{sim}}
\newcommand{\Dsim}{\vv D_\text{sim}}

\newcommand{\Wo}{\text{Wo}}
\newcommand{\Rey}{\text{Re}}

\newcommand{\set}[3]{\{#1_{#2}\}_{#2=1}^{#3}}

\newcommand{\As}{\vv A_{s}}
\newcommand{\as}{\vv a_{s}}
\newcommand{\asi}{\vv a_{s}^{(i)}}

\newcommand{\Zs}{\vv Z_{s}}
\newcommand{\zs}{\vv z_{s}}
\newcommand{\deltafct}[1]{\delta\left[#1\right]}

\newcommand{\uu}{1\hspace{-4pt}1}
\newcommand{\fsur}{\vv f}
\newcommand{\avgc}[2]{\avg{#1}_{#2}}
\newcommand{\Cov}[1]{{\cal C}(#1)}
\newcommand{\Covc}[2]{{\cal C}(#1\mid #2)}
\newcommand{\zx}{z^{(x)}}
\newcommand{\zsx}{\zs^{(x)}}
\newcommand{\zsxi}{\zs^{(x,i)}}
\newcommand{\vzx}{\vv z^{(x)} }
\newcommand{\vcx}{\vv c^{(x)} }
\newcommand{\zsur}{z_\text{sur}}
\newcommand{\vzsur}{\vv z_\text{sur}}
\newcommand{\zsurx}{\zsur^{(x)}}
\newcommand{\vzsurx}{\vzsur^{(x)}}

\newcommand{\zxp}{z^{(x')}}
\newcommand{\vcxp}{\vv c^{(x')} }

\newcommand{\Ms}{M_{s}^{\phantom{T} } }

\newcommand{\MsT}{M_{s}^{T}}
\newcommand{\tr}[1]{\text{tr}\left\{ #1 \right \}}

\newcommand{\canceled}[1]{{\cancel{#1}}}  


\section{Introduction}
At the outset of this paper stands the observation that the remarkable success of neural networks (NNs) in the computer sciences is motivating more and more studies on their application in the natural and computational sciences. Such applications could be automatized lab data analysis, or the employment as a surrogate model in  expensive many-query problems such as optimization or uncertainty quantification, see e.g. \cite{Ranftl2021b, Ranftl2022b}. Here, we desire not to employ ever larger off-the-shelf models, but to construct a simple NN and incorporate into its structure our prior knowledge of the physics. For this, we will recall two observations: First, that linear differential equations can a priori be incorporated into a Gaussian process (GP) model by proper choice of the covariance function, as was demonstrated by Albert \cite{Albert2019b} at the latest edition of this very meeting. Second comes the fact that GPs correspond to Bayesian NNs in an infinite-width limit, as demonstrated by Neal \cite{Neal1996}. Both facts together imply that the physics can be incorporated a priori also into a NN model if one chooses the neural activation functions such that they match the corresponding GP's covariance function. The aim of this work is to introduce a formal notion of this connection between physics and NNs through the lens of probability theory, and demonstrate a so derived principle for the construction of physical NNs with a simple example. It should be noted that this ansatz differs substantially from so-called "physics-informed" learning machines \cite{Karniadakis2021}, where a regularization term for the optimization is introduced but not incorporated into the structure.
\newcommand{\w}{\mathbf{\mathrm{w}}}
\renewcommand{\v}{\mathbf{\mathrm{v}}}

\section{Background and related work}

A brief outline is given of the basic literature on Gaussian proceses, (Bayesian) neural networks and the infinite-width correspondence between the two. Related work on physically inspired GPs and NNs is discussed.

\subsection{Gaussian processes}
Gaussian process regression or "Kriging" has been studied for decades \cite{OHagan1978}. They are today the arguably second-most popular class of machine learning models \cite{Rasmussen2006}.  GPs have also been popular in the closely related field of "Uncertainty Quantification" \cite{Bilionis2013, Schobi2015}.
A Gaussian process is essentially defined by its covariance function, often referred to as the kernel or in a wider sense correlation function. Thus, considerable work has been devoted towards the choice and design of such kernel functions \cite{Duvenaud2014}.

It is known that a linear operator applied to a GP again yields a GP but with a modified kernel \cite{Swiler2020}. This notion was also specialized to linear differential operators \cite{VandenBoogaart2001}, and then applied to construct kernels for physical laws expressed in terms of so-defined differential equations, e.g. divergence-free fields \cite{Jidling2017}, the Helmholtz equation \cite{Albert2019b} and the Poisson equation \cite{Dong1989}. I.e. this resulted in GPs which a priori, before training, are guaranteed to be divergence-free or obey the Helmholtz equation, respectively. The incorporation of other a priori knowledge such as boundary conditions was suggested by \cite{Graepel2003} and \cite{Gulian2022}. For an overview of more general linear constraints on GPs, see also \cite{Swiler2020}.

Abovementioned approaches for constructing physical kernels \cite{Albert2019b,VandenBoogaart2001} are fundamentally different from the popular "physics-informed GP" as popularized by \cite{Raissi2017b} and earlier introduced by \cite{Sarkka2011, Graepel2003}. The physical GP differs from the physics-informed GP in the sense that the model is not merely "informed" of the physics post-hoc through the training, but the model is constructed such that it obeys the physics a priori. The physics-informed approach consists in a GP as ansatz function and regularizing the optimization by its residual. It is known that such a physics-informed GP generalizes better than their un-regularized counterparts \cite{Karniadakis2021}. However, the physics-informed approach does not consider the form of the base kernel, the defining element of the GP, at all. Further, the physics-informed approach tolerates for the regressor to actually violate the physical requirements. The distinction of physical GPs from physics-informed GPs can so also be understood as hard constraints in contrast to weak constraints on the model.
It was also proposed to impose a GP prior on the inhomogeneity instead of the solution ansatz, and then construct the solution from the Green's function \cite{Alvarez2013, Lopez-Lopera2021}.

\subsection{Neural networks}
The arguably first-most popular class of machine learning models are NNs. Analogously to GPs and kernels, it is widely understood that the choice and design of activation functions is crucial for a NN. Many activations have been proposed, see e.g. \cite{Duvenaud2014,Apicella2021} for a recent and comprehensive overview. \cite{Ngom2021} proposed trigonometric activation functions, which by coincidence are very similar to the example that will be presented below.
The design and choice of these activations are however seldomly guided by principles and methods but rather trial and error. A small number of principled approaches through kernels, mostly relying again on the infinite-width correspondence, have been suggested \cite{Tsuchida2018, ChoSaul2009, Williams1996, Pearce2020}, cf. Sec. \ref{sec:related-work-infinite-width-correspondence}. Vice versa, the infinite-width correspondence has also been used to compute kernels for GPs corresponding to particular activations.

The term "physics-informed neural network" was popularized by \cite{RAISSI2019}, building on the earlier work of  \cite{Lagaris1998, Dissanayake1994}. Physics-informed NNs are very similar to physics-informed GPs in the sense that it serves as an ansatz function, the optimization of which is regularized by its residual.
The "physics-informed" approach has been dominating due to its simplicity, easy implementation and wide applicability, despite suffering from the usual problems of NNs and inconsistencies in the multi-objective training \cite{Rohrhofer2021}. In contrast, little attention was given to the "physics-constrained" approach, which aims at enforcing a hard constraint  in contrast to a weak constraint implied by the physics-informed approach. For this, particularly the enforcement of boundary conditions has been of special interest  \cite{Mohan2020}. Also the incorporation of symmetries has been investigated \cite{Mattheakis2019}, e.g. rotation symmetries \cite{Tsuchida2018}. Comprehensive reviews on physics-informed learning have become available recently, yet little attention has been given to the kernels and activations  \cite{Cuomo2022,Karniadakis2021}. 

\subsection{The infinite-width correspondence} \label{sec:related-work-infinite-width-correspondence}
As discovered by Neal \cite{Neal1996}, GPs and single-layer NNs are, under certain conditions and from a Bayesian point of view,  equivalent in the infinite-width limit. Thus, Neal's insight allows to apply the theory of GPs to the analysis of NNs, and vice versa.  It was later understood that Neal's theorem can be generalized to multi-layer networks \cite{Lee2018}, i.e. deep learning, and also holds after training \cite{Jacot2018} (i.e. with the posterior and not just the prior). This has led to a number of insights, e.g. that dropout regularization can be viewed as a particular prior in a Bayesian approximation \cite{Gal2016}.
Neal's theorem has also been generalized to convolutional NNs \cite{Novak2019}, and transformers or attention networks \cite{Hron2020}. The infinite-width correspondence is proven to be relevant for many real world, large scale applications.
Williams was supposedly the first to derive a GP kernel from the infinite-width limit of a single-layer NN with ReLU activations, giving the rise to what many refer to as the "neural-net induced GP" \cite{Williams1996}. William's result has also been generalized to multi-layer networks \cite{Tsuchida2018}. Others followed in deriving GP kernels from a number of activation functions \cite{ChoSaul2009,  Hazan2015}. \cite{Pearce2020} showed that various architectures and combinations of NNs correspond to combinations of GP kernels, e.g. multiplication of latent activations corresponds to a product-form of the kernel. \cite{Pearce2020} also derived a kernel corresponding to cosine functions as activation.

 
%
%

\section{Method for deriving physical activations}\label{sec:methods-approach}
\newcommand{\nn}{\mathcal{F}}
\newcommand{\lop}{\widehat{\mathcal{O}}}
\newcommand{\lopad}{\mathcal{O}^{\dagger}}
\newcommand{\exc}[1]{\big\langle #1 \big\rangle}
A formal notion of the infinite-width correspondence between GPs and NNs will be introduced, followed by procedures for the translation of linear differential constraints between GPs and NNs. For this, we will formally introduce GPs and review principles \cite{VandenBoogaart2001, Albert2019b} for the design of physical kernels. Then, we will introduce a simple single-layer neural network and analyse the limit of infinite-width, providing the infinite-width correspondence of neural networks (NNs) to Gaussian processes (GPs). We will then show how Albert's principle for GPs is translated to NNs.
We denote as $g(x)$ and $f(x)$ two learnable functions, which are governed by the same physical laws and depend on the same input,  $x \in X \subseteq \mathbb{R}$ and $g,f: X \to Y \subseteq \mathbb{R}$. For $g$ and $f$ we will assume a GP and a NN, respectively.
\subsection{Gaussian processes and physical kernels}
We assume $g$ to be a zero-mean GP,
\begin{align} \label{eq:def-GP}
    g(x) \sim \mathcal{GP}\big(0, k(x,x')\big)
\end{align}
where $k(x,x')$ is the to-be-determined covariance function or, synonymously, the kernel. An extension to non-zero mean functions should be straight forward.
We further know a priori that $g$ is governed by a physical law expressed in terms of the linear operator $\lop_x$ as
\begin{align} \label{eq:def-PDE}
    \lop_x g(x) = 0 \;,
\end{align}
where $\lop_x$ is the operator defining the physical law in dependence on the variable $x$ and acting on $g$ through $x$. This operator could be a differential operator defining a differential equation, e.g. the Hamilton operator or more specifically the operator that defines the Helmholtz equation:
\begin{align}
    \lop_x \; \hat{=} \; \nabla^2 + \nu^2 \;,
\end{align}
with some constant wave number $\nu$. Other examples would be $\lop \hat{=} \nabla^2$, then this would be the Laplace equation, and would imply that $g$ obeys this Laplace equation. It would also imply that all measurements $y$, i.e. observed data of $g(x)$, namely $y = g(x) + \eta$, would obey this Laplace equation up to the measurement noise $\eta$. If the noise is independent of the input and the operator is linear, then $\lop y = \eta$. This latter fact could hypothetically inform us on the choice of the likelihood, but that is not the matter of this work.

Let us now find a kernel $k(x,x')$ such that \eqref{eq:def-PDE} is indeed fulfilled.
It was shown by van den Boogaart \cite{VandenBoogaart2001} that if \eqref{eq:def-PDE} and \eqref{eq:def-GP} hold simultaneously, that this is equivalent to
\begin{align} \label{eq:def-boogaart}
    \lop_x \lop_{x'} \;k(x,x') \Big\rvert_{x=x'} & \stackrel{!}{=} 0 \;.
\end{align}
That means, that if we find a kernel $k$ such that \eqref{eq:def-boogaart} is fulfilled, than the Gaussian process \eqref{eq:def-GP} defined by that kernel $k$ also obeys \eqref{eq:def-PDE} and vice versa. It is nothing else than that we demand the GP to fulfill the PDE exactly, and the variance or uncertainty of the GP be zero at any particular probing location aside from a nugget that would model a measurement error.
Note that Boogaart showed this equivalence also for non-zero mean functions.
Mercer's theorem suggests a principled approach to solving \eqref{eq:def-boogaart}, in that it states that a kernel can be represented by a sum of products of basis functions $\phi$ \cite{schaback_wendland_2006},
\begin{align} \label{eq:def-mercer}
    k(x,x') &= \lim_{N\to\infty} \sum_{i,j}^{P} \phi_i(x) M_{ij} \phi_j(x')
\end{align}
with some suitable prior covariance matrix $M$ for the function weights (not shown). It was argued by Albert \cite{Albert2019b} that Mercer's kernel can be built by considering the basis functions $\phi$ to be fundamental solutions to the differential equations \eqref{eq:def-PDE}. 
Going from right to left in \eqref{eq:def-mercer} with a given basis may also be understood as "the kernel trick" \cite{Rasmussen2006}, where inner products of large bases are substituted with the corresponding kernel in order to avoid the computation of large matrices. We will later make an attempt at something like an "inverse" of this kernel trick, i.e. going from left to right and substituting the kernel through its basis function representation.
Substitution of \eqref{eq:def-mercer} into \eqref{eq:def-boogaart} yields
\begin{align}
    \lim_{N\to\infty} \sum_{i,j}^{N} \underbrace{\Big(\lop_x \phi_i(x) \Big)}_{=\psi_i(x)} M_{ij} \underbrace{\Big(\lop_{x'} \phi_j(x') \Big)}_{=\psi_j(x')} &= 0 \;,
\end{align}
where linearity was used and that the operators only act on one term in the product each. One may argue that the operator acting upon a basis function $\phi$ should give rise to a new function $\psi$. It would be interesting under which circumstances this defines a new basis.

\subsection{Neural networks and the infinite-width correspondence}
Let us consider 
a  NN $f$ with a \textit{single} hidden layer with $N$ neurons with non-linear but bounded activation functions $h$,
\begin{align} \label{eq:def-nn}
    f(x) = \sum_{k=1}^N \v_k h_k \Big( \w_k \cdot x + a_k\Big) + b \;, 
\end{align}
where the index $k = 1,...,N$ sums over the $N$ neurons in the single hidden layer. $\w_k$ are the input-to-latent weights,  $a_k$ and $b$ are bias terms and $\v_k$ are the latent-to-output weights. Generalization to multiple features is straight-forward through adding another index to the weights.
In the infinite-width limit, $N\to\infty$, \eqref{eq:def-nn} converges to a Gaussian process by the Central Limit Theorem if the activation functions $h_k$ are all identical and bounded \cite{Neal1996}.
Let $b$ and all $\v_k$'s have zero-mean, independent Gaussian priors with variances $\sigma_b^2$ and $\sigma_{\v_k}^2$ respectively. 
Let $\exc{\cdot}$ denote the expectation value with respect to all weights. Then $\exc{f(x) \cdot f(x')}$ is the NN's output's covariance. A specific NN is equivalent to a specific GP with kernel $k(x,x')$ if the NN's covariance is equal to the kernel,
\begin{align} \label{eq:infinite-width-correspondence}
    k(x,x')  & \stackrel{!}{=} \exc{f(x) f(x')} \notag \\
    &= \sigma_b^2 + \lim_{N\to\infty} \sum_{k=1}^{N} \sigma_{\v_k}^2 \exc{h_k(x;{\w_k}) \cdot h_k(x'; {\w_k})} \;, \\
    &= \sigma_b^2 + A^2 \exc{h(x;{\w}) \cdot h(x'; {\w})} \;, \notag
\end{align}
where we used the linearity of the expectation in the first line. In the second line, the limit has been carried out under the assumption of identical neurons $h_k = h$ after fixing $\sigma_{\v_k} = A/\sqrt{N}$ for some fixed $A$, as argued by Neal \cite[ch.2]{Neal1996} or Williams \cite{Williams1996}.
This relationship has been used to compute GP kernels for given neural activation functions, e.g. a sigmoidal activation \cite{Williams1996}.
Note that it is crucial that the prior reflects that the hidden units are independent, as otherwise the Central Limit Theorem no longer holds.
In order to design a NN that a priori obeys the physics, it would be already sufficient to find neural activations $h$ such that \eqref{eq:infinite-width-correspondence} holds for a particular GP kernel fulfilling \eqref{eq:def-boogaart}. We may however observe other possibly useful relationships.
\subsection{Physical neural activation functions}
We may now insert the Mercer representation of the kernel \eqref{eq:def-mercer}  on the left hand side of the infinite-width correspondence \eqref{eq:infinite-width-correspondence}

\begin{align}
     \lim_{N\to\infty} \sum_{i,j}^{N} \phi_i(x) M_{ij} \phi_j(x')    &= \sigma_b^2 + \lim_{N\to\infty} \sum_{k=1}^{N} \sigma_{\v_k}^2 \exc{h_k(x;{\w_k}) \cdot h_k(x'; {\w_k})} \;.
\end{align}
Let the biases be known and fixed, i.e. $\sigma_b = 0$, then one solution to this equation is
\begin{subequations}\label{eq:fundamental_conditions}
\begin{align}
    M_{ij} &= \sigma_{\v_i} \delta_{ij} \;, \label{eq:fundamental_conditions_A}\\
    p({\w_k}) &= \delta(\widehat{\w}_k) \;,   \label{eq:fundamental_conditions_B}\\
    h_m(x;{\w_m}) &= \phi_m(x)  \;. \tag{10c}\label{eq:fundamental_conditions_C}
\end{align}
\end{subequations}
%
The input-to-latent weights $\w_m$ now correspond to the implicit parametrizations of the GP basis functions $\phi_m$.
That would mean, that the functions defining the Mercer representation of a GP-kernel can be particularly useful for finding the activation functions of the corresponding equivalent NN in the infinite width limit if the weights' prior is chosen according to the Mercer representation.
For some cases, there will be analytical solutions.
We may now coerce the GP to obey some physical law $\lop$ and find
\begin{align} \label{eq:boogaart-neal-combo-pulled-through}
     \lim_{P\to\infty} \sum_{i=1}^{N}  \sum_{j=1}^{N} \psi_i(x) M_{ij} \psi_j(x') & {=}  \lim_{N\to\infty} \sum_{k=1}^{N} \exc{ \underbrace{\lop_x h_k(x;{\mathrm w})}_{=\tilde{h}_k(x;{\mathrm w})} \cdot \underbrace{\lop_{x'} h_k(x'; {\mathrm w})}_{=\tilde{h}_k(x;{\mathrm w})}}_{\mathrm w} \;.
\end{align}
The application of the differential operator to activation functions $h$ yields new functions $\tilde{h}$.
By finding the Mercer representation of a kernel, which coerces a GP to obey a physical law, we can find corresponding neural activation functions which are capable to force a NN to obey the same physical law, too.

\subsection{Training}\label{subsec:training}
Given pairs of input-output data $D= \{x^{(d)}, y^{(d)}\}_{d=1}^{N_d}$, we ought to choose a criterion or loss function for the optimization. The optimal parameters $\w^\ast$ are defined by the loss function, here 
\newcommand\norm[1]{\left\lVert#1\right\rVert}
\begin{align}
    &\w ^{\ast}   = \arg \min_{\mathbf{w}} \Bigg[\sum_{d=1}^{N_d} \Big(y^{(d)} - f(x^{(d)}) \Big)^2 + \lambda \sum_{p=1}^{N_p} \norm{\Big( \lop_{x^{}} f(x^{})\Big)\Big|_{x=x^{(p)}}} \Bigg] \;, \label{eq:training}
\end{align}
which consists of a 'data loss', i.e. the sum of squared errors, and a 'physics-loss', i.e. the residual of \eqref{eq:def-PDE} if network $f$ is substituted as ansatz function, and a tuneable regularization parameter $\lambda$. The second term on the right hand side is to be understood as the differential operator applied to the function $f$, which is then evaluated at $N_p$ pivot points $x^{(p)}$. A suitable norm $\norm{\cdot}$ is to be chosen.
For the proposed physics-constrained network, the physics-loss vanishes and $\lambda$ can be arbitrary. For a vanilla neural network, $\lambda$ would be set to zero. For a physics-informed network (see also \cite{RAISSI2019}), a finite $\lambda$ is chosen. The outcome is generally highly sensitive to the choice of $\lambda$, hence usually a hyperparameter optimization must be conducted.
%
%
%
%
\section{Numerical example}
\begin{figure}[htbp]
\begin{subfigure}[h]{0.49\textwidth}
    \centering
    \includegraphics[width=1\textwidth]{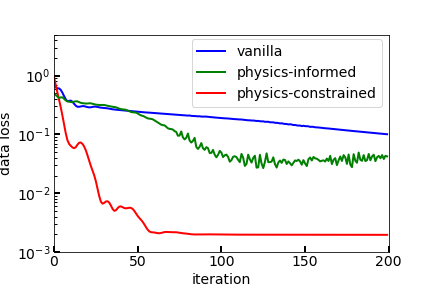}
    \label{fig:examples-solutions}
\end{subfigure}
\begin{subfigure}[h]{0.49\textwidth}
    \centering
    \includegraphics[width=1\textwidth]{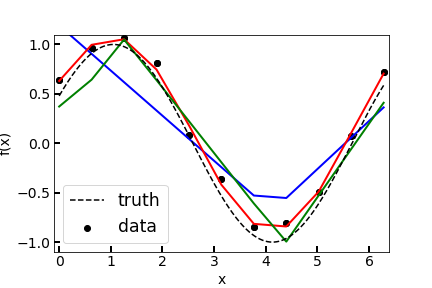}
    \label{fig:convergence}
    \end{subfigure}
    \caption{Plot of convergence (left) and corresponding solutions (right). For the visualization only, only the data loss is shown since the physics loss vanishes in the physics-constrained case.  'Vanilla' (blue) refers to an ordinary NN with ReLU activations. 'Physics-informed' (green) refers to the same NN but with additional training regularization by the physics-loss, \eqref{eq:training}. 'Physics-constrained' (red) refers to again the same NN but with physics-based activations  as proposed in this paper. }\label{fig:plot}
\end{figure}

The approach shall be illustrated with the simple example of the one-dimensional homogeneous Helmholtz equation, i.e.
\begin{align} \label{eq:helmholtz-equation}
 \Big( \frac{\partial^2}{\partial x^2} + \nu^2 \Big) f(x) = 0 \;.
\end{align}
The kernel derived from our first principles for the Helmholtz equation, referred to as the (1D-) Helmholtz kernel from hereon, has the following simple form \cite{Albert2019}:
\begin{align}
k(x,x') = \cos\big(\alpha(x-x')\big) \;.
\end{align}
It should be noted that $\cos\big(\alpha(x-y)\big)  = \sin(\alpha x)\sin(\alpha y) + \sin(\alpha x+\Delta)\sin(\alpha y+\Delta)$ with phase $\Delta $. Following \eqref{eq:infinite-width-correspondence} and under \eqref{eq:fundamental_conditions} we may then choose sinusoidal activation functions
\begin{align}
h(x) = \sin{(x)} \;,
\end{align}
which correspondingly will be referred to as the (1D-) Helmholtz activation. The 'frequency' $\alpha$ and 'phase' $\Delta$ are then to be learnt by the NN as weights $\mathrm{w_k}$ and biases $a_k$.
We now proceed to illustrate the example numerically. The implementation is remarkably simple, and with high-level libraries such as PyTorch amounts circa to the re-writing of one line of code in a vanilla NN. For the following experiments, 11 noisy observations at equidistant pivot points from the fundamental solution to \eqref{eq:helmholtz-equation} have been generated with a noise level of $20\%$, a frequency of $\omega = 0.51$ and $\phi=0.50001$. The role of the choice of pivot points is not of primary concern here. 
As optimizer we choose ADAM \cite{Kingma2015a}  with a learning rate of $0.02$. In the case of the physics-informed NN, the weight for the physics-loss was set to $\lambda = 0.1$. We note that experiments with stochastic gradient descent (various learning rates between $0.001$ and $1.0$) usually did not lead to sufficient convergence of the  physics-informed NN (ReLU activations and  training regularization, $\lambda \neq 0$) and the vanilla NN (ReLU activations and no training regularization, $\lambda = 0$) within a few thousand iterations, while the here constructed physical NN still converged (albeit slower). 
Fig. \ref{fig:plot} (right) presents the learned solutions to the Helmholtz equation \eqref{eq:helmholtz-equation}, and Fig. \ref{fig:plot} (left) shows convergence plots. For reference, our physically-constructed NN is compared to a merely physics-informed NN (i.e. regularized training as in discussed in Sec. \ref{subsec:training}) with ReLU activations as well as a plain vanilla NN with ReLU activations. Unsurprisingly, the NN that has the physics incorporated into its structure and design vastly outperforms its merely informed or non-informed counterparts.

\section{Discussion}
In the above presented example, it turns out that the obtained activation function is also the fundamental solution to the differential equation \eqref{eq:def-PDE}. Since the kernel may be built from fundamental solutions as basis functions $\phi$ in Mercer's representation \eqref{eq:def-mercer}, it seems that under the conditions \eqref{eq:fundamental_conditions} also the neural activations $h$ should be proportional to those fundamental solutions.
It may seem sensible to construct the solution from a combination of fundamental solutions, as it amounts to for this example. It should also be observed that our so constructed NN also has the form of a Fourier series. Curiously, at the outset of the modern day perspective on regression through NNs, we have arrived at the conclusion that the proper physical NN assumes the simple forms that have been known all along. While this could be perceived as a circular journey, it seems reasonable that the result should be consistent with earlier perspectives on the same problem.

The here discussed connection can, in theory, be exploited to obtain a principled catalogue of equivalences between physical laws and their corresponding kernel functions as well as activation functions. 
There are at least three conceivable procedures for finding solutions to \eqref{eq:boogaart-neal-combo-pulled-through}.
One procedure would be to find matching examples in a forward manner, similar to \cite{Williams1996}, by defining neural activations by trial and error until they match a particular kernel. Any of the several representations from above may be used for that.
This can be guesswork, if not less effective or necessarily less efficient. A more systematic approach would be through Mercer's kernel representation, e.g. with fundamental solutions as basis functions as argued before, implying an "inverse kernel trick". For many physical laws, this might be possible numerically only, if at all, which could be crucial for adoption of the procedure to more complex physical laws. It is also argued that it is possible to "skip" the construction of the kernel and directly find the activation function by demanding the right hand side of \eqref{eq:boogaart-neal-combo-pulled-through} be zero. It should further be noted that Boogaart \cite{VandenBoogaart2001} also displayed equivalent formulations of \eqref{eq:def-boogaart} and gives "four methods, too old to be found in the books I read". These should be useful for constructing physical activations, too.

A limitation of the approach consists in the fact that our basic assumption were linear operators. I.e., non-linear equations cannot be treated in this way directly in the sense that non-linear operators applied to GPs do not yield GPs anymore. Hence \eqref{eq:def-boogaart} for constructing physical kernels does not hold anymore. As a consequence, no physical activations can be derived in the infinite-width correspondence either. A mitigation of this problem could lie in the linearization of the non-linear equations in order to retain the original assumptions. For highly non-linear effects, such an approximation may however be problematic.
Another limitation is that in reality NNs always have finite width. It remains unclear how wide a NN really needs to be for behaving approximately like a GP in this context.

We seek also to address limitations of the experimental study. For one, no optimization of architecture details and hyperparameters has been carried out. 'Vanilla' NNs should be well capable of learning the Helmholtz equation if enough neurons are added, and if the optimization scheme is chosen appropriately. Particularly the performance of the vanilla physics-informed NN depends strongly on the choice of the  weighting hyperparameter $\lambda$ in the multi-objective loss. It is well possible that, if chosen appropriately, such a NN might outperform. How an appropriate $\lambda$ can be found efficiently however is an open problem to date. The claim of this paper is not that the here constructed physical neural activation functions generally outperform physics-informed NNs, although it does in the above presented example. This work merely points out that the connection between GPs and NNs can be exploited for the principled design of physical neural activations based on linear differential equations. In contrast to a physics-informed NN, the physics-loss in \eqref{eq:training} rhs then vanishes a priori, and hence the tuning of the highly sensitive hyperparameter $\lambda$ becomes obsolete.

Finally, many open questions remain, particularly also in anticipation of the various generalizations and other applications of the infinite-width correspondence, cf. Sec. \ref{sec:related-work-infinite-width-correspondence}. Can the approach be extended to deep NNs or even more complex architectures like LSTMs, Transformers or Graph Neural Networks? Do the assumptions still hold during training? A heuristic idea would be to simply replace the activations in the last hidden layer of a deep NN with physical activations. How can we include symmetries,  boundary or initial conditions systematically, and are Green's functions a viable path for that? Can the approach be beneficial for differential equations without known fundamental solutions?

\section*{Acknowledgments}
The author would like to thank Wolfgang von der Linden, Thomas Pock, Christopher Albert and Robert Peharz for their helpful support and comments and the MaxEnt 2022 community for valuable discussions. The author was funded by Graz University of Technology (TUG) through the LEAD Project "Mechanics, Modeling, and Simulation of Aortic Dissection" (biomechaorta.tugraz.at) and  supported by GCCE: Graz Center of Computational Engineering.

\bibliographystyle{unsrt}  
\bibliography{references}  
 
\end{document}